\documentclass[twocolumn, 10pt, final] {IEEEtran}
\usepackage{cite}
\usepackage{amsmath,amssymb,amsfonts}
\usepackage{algorithmic}
\usepackage{graphicx}
\usepackage{textcomp}
\usepackage{wrapfig}

\usepackage{color, colortbl}
\definecolor{lightgray}{rgb}{0.83, 0.83, 0.83}
\definecolor{subsectioncolor}{rgb}{0.0, 0.0, 0.0}
\usepackage{tikz}
\usepackage{hyperref}
\usepackage{algorithm,algorithmic}
\usepackage[flushleft]{threeparttable}
\usepackage{hyperref}
\usepackage{textcomp}
\newcommand{\ins}{\text{\tiny{ins}}}

\def\BibTeX{{\rm B\kern-.05em{\sc i\kern-.025em b}\kern-.08em
    T\kern-.1667em\lower.7ex\hbox{E}\kern-.125emX}}

\definecolor{abstractbg}{rgb}{0.89804,0.94510,0.83137}
\begin{document}
\title{MAINS: A Magnetic Field Aided Inertial Navigation System for Indoor Positioning}
\author{Chuan Huang, \IEEEmembership{Student Member, IEEE}, Gustaf Hendeby, \IEEEmembership{Senior Member, IEEE},  Hassen Fourati, \IEEEmembership{Senior Member, IEEE},  Christophe Prieur, \IEEEmembership{Fellow, IEEE}, and Isaac Skog, \IEEEmembership{Senior Member, IEEE}
\thanks{This work has been funded by the Swedish Research Council (Vetenskapsrådet) project 2020-04253 "Tensor-field based localization". }
\thanks{Chuan Huang and Gustaf Hendeby are with Dept. of Electrical Engineering, Linköping University, (e-mail: chuan.huang@liu.se; gustaf.hendeby@liu.se).}
\thanks{Hassen Fourati and Christophe Prieur are with the GIPSA-Lab, CNRS, Inria, Grenoble
INP, University Grenoble Alpes, 38000 Grenoble, France (e-mail: hassen.fourati@gipsa-lab.fr; christophe.prieur@gipsa-lab.fr)}
\thanks{Isaac Skog is with 
Dept. of Electrical Engineering, Uppsala University, Uppsala, Sweden, and Dept. of Electrical Engineering
Linköping University, Linköping, Sweden, and the Div. of Underwater Technology, Swedish Defence Research Agency (FOI), Kista, Sweden (e-mail: isaac.skog@angstrom.uu.se).}}

\maketitle

\begin{abstract}
 A Magnetic field Aided Inertial Navigation System (MAINS) for indoor navigation is proposed in this paper. MAINS leverages an array of magnetometers to measure spatial variations in the magnetic field, which are then used to estimate the displacement and orientation changes of the system, thereby aiding the inertial navigation system (INS). Experiments show that MAINS significantly outperforms the stand-alone INS, demonstrating a remarkable two orders of magnitude reduction in position error. Furthermore, when compared to the state-of-the-art magnetic-field-aided navigation approach, the proposed method exhibits slightly improved horizontal position accuracy. On the other hand, it has noticeably larger vertical error on datasets with large magnetic field variations. However, one of the main advantages of MAINS compared to the state-of-the-art is that it enables flexible sensor configurations. The experimental results show that the position error after 2 minutes of navigation in most cases is less than 3 meters when using an array of 30 magnetometers. Thus, the proposed navigation solution has the potential to solve one of the key challenges faced with current magnetic-field simultaneous localization and mapping (SLAM) solutions — the very limited allowable length of the exploration phase during which unvisited areas are mapped.
\end{abstract}


\section{Introduction}
\label{sec:introduction}
\begin{figure}[htb!]
  \centering
     \includegraphics[width=\columnwidth]{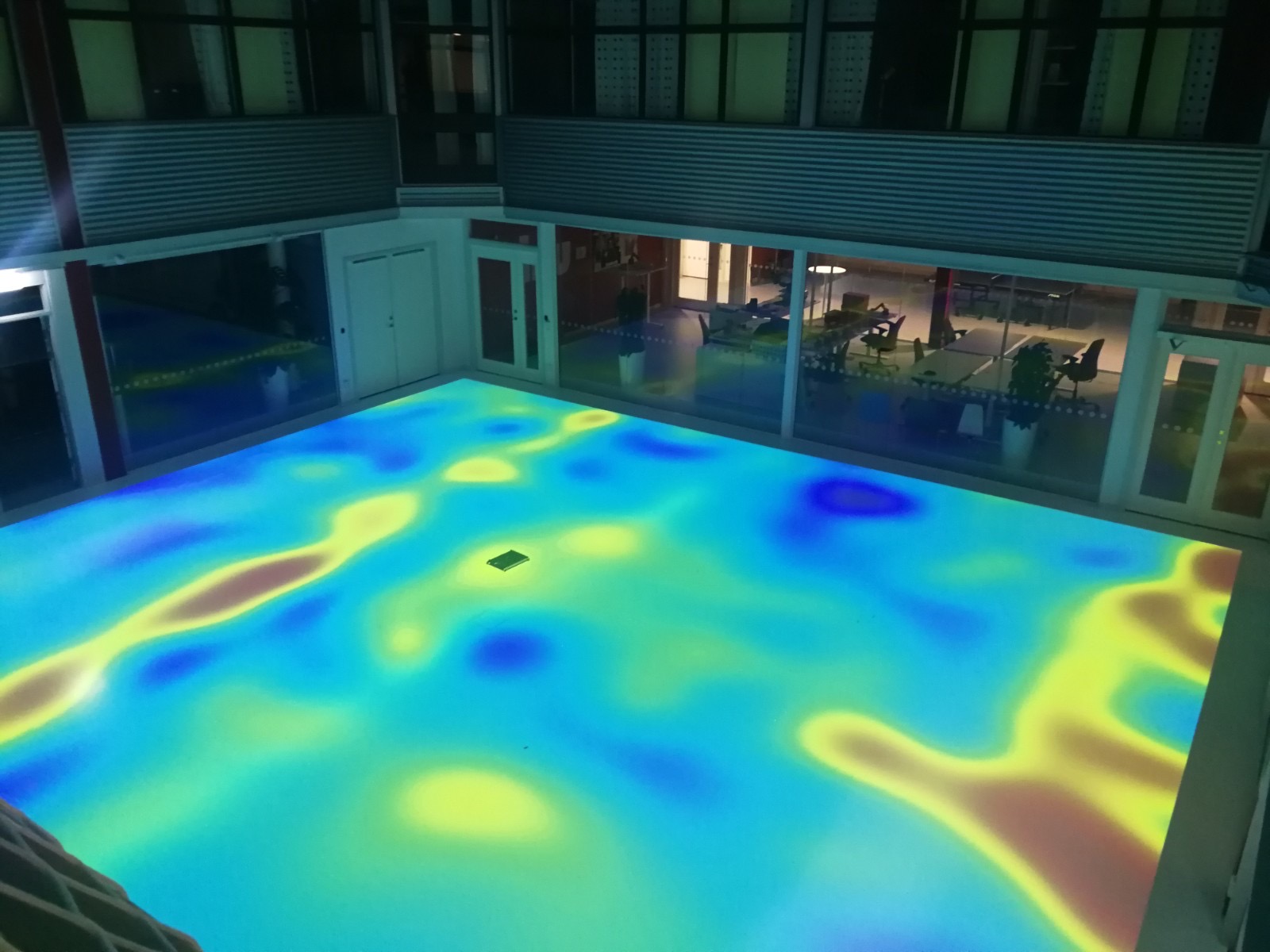}
      \caption{Illustration of the magnetic-field magnitude variations inside a building. The field near the floor was measured with a magnetometer, whose location was tracked by camera-based tracking systems. The field measurement was then interpolated, and the field magnitude was projected on the floor.}
      \label{F:field}
  \end{figure}

The outdoor magnetic field is omnipresent, relatively stable, and almost homogenous. Due to these properties, it has been used in navigation for a long time, where the magnetic field is mainly used as a heading reference to correct errors from integrating noisy gyroscope measurements\cite{chesneau2018magneto}. However, those techniques cannot be applied without modifications for indoor applications because the indoor magnetic field is not homogenous. An example of the variations in the magnitude of the magnetic field inside a building is shown in Fig.~\ref{F:field}. The correlation between the position and the magnetic field can be seen. Therefore, the inhomogeneous magnetic field can be used as a reliable source for localization in Global Navigation Satellite System (GNSS) denied environments, such as indoors or underwater\cite{skog2021magnetic}.
Indeed, recent years have witnessed many successful applications in magnetic-field-based positioning, among which the magnetic-field-based SLAM has turned out to be a promising approach\cite{Kok2018MagSlam, Viset2022EKF, manon2021MagneticField}. It enables the user to construct a magnetic field map while navigating and having drift-free positioning, provided revisiting the same region is possible. However, this technology relies heavily on the precision of the used odometric information. Otherwise, the position drift can be significant, making it challenging to reliably recognize the visited place and complete ``loop closure''\cite{thrun2005probabilistic}. For instance, when an inertial navigation system with low-cost inertial sensors is used for doing the odometry, the error growth rate is typically on the order of 10 meters per minute \cite{Nilsson2016Inertial}, which means the system needs to revisit the same region within a minute to prevent the position drift from becoming too large to complete loop closure. Therefore, the permissible length of the exploration phases where new areas are mapped is extremely limited when using low-cost inertial sensors. Hence, to increase the usability of current magnetic-field-based SLAM solutions we need robust odometry techniques that have a low position drift rate.

With this limitation in mind, the concept of combining inertial measurements and distributed magnetometry has been proposed and realized~\cite{Vissiere2007magnetometers}. An array of magnetometers enables the calculation of the gradients of the magnetic field, from which the body velocity can be estimated. On the other hand, the authors in \cite{skog2021magnetic} adopted a model-based approach, treating pose changes in subsequent timestamps as the parameters of the model to be estimated. The estimated pose change from the magnetic field measurements can be used to aid an inertial navigation system to reduce the position growth rate. In this paper, we present a method for tightly integrated magnetic-field-aided inertial navigation. The resulting navigation system has, compared to a pure inertial navigation system, a significantly reduced error growth rate. Hence, the proposed navigation method has the potential to greatly extend the allowable length of the exploration phases in magnetic-field-based SLAM systems.

\subsection{Related Work}
Numerous methods for magnetic-field-based indoor positioning and navigation have been proposed. Current solutions are roughly categorized into three types: fingerprint-based methods, magnetic field SLAM, and magnetic field odometry. Fingerprint-based methods\cite{Shi2023Pedestrain, Liu2023MagLoc, Pasku2017Magnetic} generally rely on a premeasured magnetic field map to work, which greatly limits its usability. Therefore, we mainly discuss the latter two types, which are not constrained by a prior map.

One of the first 2D magnetic field SLAM methods was proposed in \cite{Robertson2013Simutaneous}, where the magnetic field strength map was constructed in hierarchical hexagonal tiles of different bin sizes. To generalize the motions to 3D and cope with the complexity of representing the map, the authors in \cite{Kok2018MagSlam} selected a reduced-rank Gaussian process to represent the magnetic field map and used a Rao-blackwellized particle filter to estimate both the map and the location of the system. Later, in \cite{manon2021MagneticField}, the authors successfully achieved drift-free positioning using foot-mounted sensors with similar techniques. To achieve real-time processing, they used an EKF filter to update the magnetic field map based on its gradients\cite{Viset2022EKF}. Contrary to the ``stochastic'' loop closure mechanism in \cite{Kok2018MagSlam, manon2021MagneticField, Viset2022EKF}, where it reduces the uncertainty of the map in the revisited region, a ``deterministic'' one was employed in \cite{pavlasek2023magnetic}, where attitude-invariant magnetic field information was used to detect loop closure and constraints on position estimates were formed. The aforementioned magnetic field SLAM solutions can achieve drift-free long-term large-scale positioning as long as frequent loop closure is possible.

Magnetic field odometry, on the other hand, is a lightweight solution to provide odometric information with a magnetometer array. It does not construct a magnetic field map but provides odometric information based on local magnetic field properties. The seminal work \cite{Vissiere2007magnetometers} derived an equation that relates the body velocity and the gradient of the magnetic field, which can be calculated from the measurements from a set of spatially distributed magnetometers. Later, in \cite{dorveaux2011combining, dorveaux2011presentation}, the authors proposed an observer to estimate body velocity, proved its convergence, and showcased its usability for indoor localization.  In subsequent works \cite{chesneau2016motion, Chesneau2017ImprovingMagnetoInertial}, the authors incorporated inertial sensor biases and magnetic disturbance in the model and designed a filter based on the error-state Kalman filter (ESKF). To address the issue of the noisy magnetic field gradient, the authors in \cite{zmitri2019improving, zmitri2020magnetic} derived a differential equation for high-order derivatives of the magnetic field. They then developed a filtering algorithm comprising a primary filter which is used to estimate the gradient of the field. Later, the same authors \cite{zmitri2020LSTM, zmitri2022BILSTM} proposed an AI-based solution where a Long Short-Term Memory (LSTM) model is used to create a pseudo-measurement of the inertial velocity of the target. The pseudo-measurement is used to handle adversarial situations where the states’ observability is affected by the low gradient of the field and/or the target's velocity close to zero. The methods developed have demonstrated promising prospects in magnetic field odometry. However, they are susceptible to noise when computing the gradient, and observability issues arise from a weak magnetic field gradient or low target speed. To address this problem, \cite{ZHANG2023MagODO} matches the waveforms from a pair of magnetometers in a sliding time window to reduce the influence of the temporary disappearance of the magnetic field gradient. Experiments show that the proposed method in \cite{ZHANG2023MagODO} performs similarly to wheel odometry in magnetic-rich environments.

In a more recent work \cite{skog2018magnetic}, a polynomial model was proposed to describe the local magnetic field and to develop a magnetic field odometry method. Presented experiential results showed that the model-based odometry approach can give a higher accuracy at low signal-to-noise ratios, compared to approaches in \cite{Vissiere2007magnetometers}. The model-based odometry approach was further explored in \cite{skog2021magnetic}, where it was used to estimate both the translation and orientation change of the array. In the subsequent work \cite{Chuan2022Magnetic}, the authors included the magnetic field model in the state-space description of an INS system and developed a tightly-integrated magnetic-field-aided INS. Simulation results showed that it has a much slower drift rate than stand-alone INS.

\subsection{Contributions}
The contribution of this work is two-fold. Firstly, it extends the groundwork laid out in \cite{Chuan2022Magnetic} by providing a thorough derivation and a comprehensive exposition of the proposed algorithm. Additionally, the performance of the proposed algorithm using real-world data was assessed and benchmarked against the state-of-the-art. Secondly, we have made the datasets used in our experiments and the source code for the proposed algorithm MAINS publicly available, in the hope that it will facilitate further research within the area of magnetic-field-based positioning. Both the datasets and the source code are available at \url{https://github.com/Huang-Chuan/MAINSvsMAGEKF}.

\section{System Modeling}
\begin{figure}[tb!]
    \centering
    \includegraphics[width=\columnwidth]{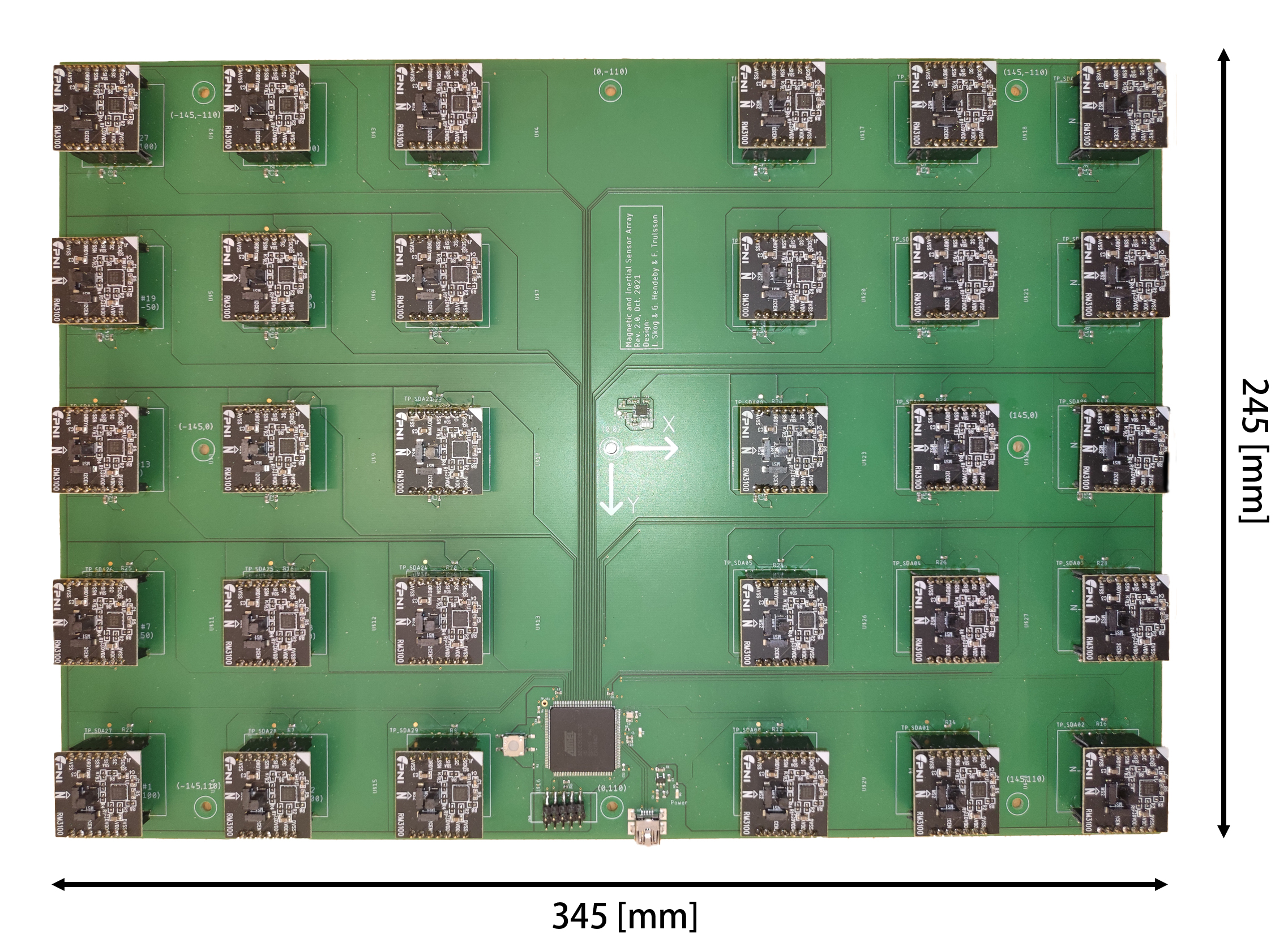}
    \caption{The sensor board used in the experiment. It has 30 PNI \href{https://www.pnicorp.com/rm3100/}{RM3100} magnetometers and an Osmium MIMU 4844 IMU mounted on the bottom side.}
    \label{F: sensor board}
\end{figure}

Consider the problem of estimating the position and orientation of the sensor platform in Fig. \ref{F: sensor board}, which consists of an inertial measurement unit (IMU) and an array of 30 magnetometers. To that end, a state-space model will be presented to realize a tightly-integrated magnetic-field aided inertial navigation system (INS). 

\subsection{Inertial Navigation Equations}
Let the INS navigation state $x^{\ins}_k$, the inertial measurements $\tilde{u}_k$, and the process noise $w^{\ins}_k$ be defined as 
\begin{equation}
        x^{\text{\tiny{ins}}}_k \triangleq \begin{bmatrix}
    p_k^n \\
    v_k^n \\
    q_k\\
    o^{a}_k\\
    o^{\omega}_k\\
    \end{bmatrix}, \quad
    \tilde{u}_k \triangleq \begin{bmatrix}
        \tilde{s}_k\\
        \tilde{\omega}_k
    \end{bmatrix}, \quad \text{and} \quad
    w^{\ins}_k \triangleq \begin{bmatrix}
        w_k^a \\
        w_k^{\omega} \\
        w_k^{o^{a}} \\
         w_k^{o^{\omega}}
    \end{bmatrix},
\end{equation}
respectively. Here, $p^n_{k} \in \mathbb{R}^3$, $v^n_{k}\in \mathbb{R}^3$, and $q_{k}\in \mathbb{H}$ denote the position, velocity, and orientation (parameterized as a unit quaternion) at time $k$, respectively. The superscript $n$ indicates that the vector is represented in the navigation frame. Moreover, $o^{a}_k \in  \mathbb{R}^3 $ and $o^{\omega}_k \in \mathbb{R}^3$ denote the accelerometer and gyroscope bias, respectively. Further, $\tilde{s}_k$ and $\tilde{\omega}_k$ denote the accelerometer and gyroscope measurements, respectively. Lastly, $w_k^a$ and $w_k^\omega$ denote the accelerometer and gyroscope measurement noise, respectively, and $w_k^{o^a}$ and $w_k^{o^\omega}$ denote the random walk process noise for the accelerometer and gyroscope biases, respectively. For an INS that uses low-cost sensors and moves at moderate velocities such that the effects of the transport rate, earth rotation, etc., can be neglected, the navigation equations are given by~\cite{Joan2017Quaternion}
\begin{subequations}
\label{eq: navigation equations}
\begin{align}
x^{\ins}_{k+1} & = f^{\ins}(x^{\ins}_k,\tilde{u}_k, w^{\ins}_k),
\end{align}
where  
\begin{equation}
    f^{\ins}(x^{\text{\tiny{ins}}}_k,\tilde{u}_k, w^{\text{\tiny{ins}}}_k) \!=\!\begin{bmatrix}
p_{k}^n + v_{k}^n T_s + (R^n_{b_k}s_k + g^n) \frac{T_s^2}{2} \\
v_{k}^n + (R^n_{b_k}s_k + g^n) T_s\\
q_k \otimes \text{exp}_q\left( \omega_k T_s \right)\\
o^{a}_k + w_k^{o^{a}}\\
o^{\omega}_k + w_k^{o^{\omega}}\\
\end{bmatrix},
\end{equation}
and
\begin{align}
    s_k&= \tilde{s}_k - o_k^{a} - w_k^{a},\\
    \omega_k&= \tilde{\omega}_k- o_k^{\omega} - w_k^{\omega}.
\end{align}
\end{subequations}
Here, the subscript $b_k$ denotes the body frame at time $k$, and $R^n_{b_k} \in \mathbb{SO}(3)$ denotes the rotation matrix that rotates a vector from the $b_k$-frame to the $n$-frame. Further, $T_s$ denotes the sampling period. Moreover, $s_k \in\mathbb{R}^{3} $ and $\omega_k \in \mathbb{R}^{3}$ denote the specific force and angular velocity, respectively. The vector $g^n \in \mathbb{R}^3$ denotes the local gravity. Furthermore, $\otimes$ denotes quaternion multiplication, and $\text{exp}_q(\cdot)$ is the operator that maps an axis-angle to a quaternion. Lastly, $w^{\ins}_k$ is modeled as a zero-mean white Gaussian noise process with covariance matrix $\Sigma_{w^{\ins}_k} = \text{blkdiag}(\Sigma_{a},\Sigma_{\omega},\Sigma_{o^{a}},\Sigma_{o^{\omega}})$, where $\Sigma_{(\cdot)}$ denotes the covariance matrix of the corresponding noise component and blkdiag($\cdot$) is an operator that creates a block diagnal matrix.

\subsection{Magnetic Field Modeling}
The magnetic field is a three-dimensional vector field whose properties are described by Maxwell's equations. Let $M(r;\mu)$ be a model of the magnetic field at the location $r\in \mathbb{R}^3$, parameterized by the parameter $\mu$. When there is no free current in the space $\Omega$, $M(r;\mu)$ should fulfill
\begin{subequations}
\label{eq: maxwell's equation}
\begin{align}
    \nabla_r \times M(r;\mu) &= 0, \label{eq: curl free}\\
    \nabla_r \cdot M(r;\mu) &= 0,  \label{eq: div free}
\end{align}
\end{subequations}
for all $r\in \Omega$\cite{jackson1999Classical}. 

Equation \eqref{eq: curl free} guarantees that there exists a scalar potential function $\phi(r;\mu)$ whose gradient is $M(r;\mu)$, i.e.,
\begin{equation}
    M(r;\mu) = \nabla_r \phi(r;\mu).
\end{equation}
Let $\phi(r;\mu)$ be a polynomial of order $(l+1)$ in the $r$ component written as
\begin{equation}
    \label{eq: potential}
    \phi(r;\mu)=h(r)^{\top}\mu + c.
\end{equation}
Here $h(r)$ is a vector whose elements are given by the product $r_x^i r_y^j r_z^k$ for $\forall i,j,k \in \mathbb{N}$, subject to $1 \leq i +j + k \leq m$, $m = 1,2,\cdots, l+1$. Further, $\mu \in \mathbb{R}^L$ is a column vector of dimension $L=(l+4)(l+3)(l+2)/6 - 1$ and $c$ is an arbitrary constant that does not affect the gradient of the potential function. Let $\Gamma(r)=\nabla_r h(r)^{\top}$, then $M(r;\mu)$ can be written as 
\begin{equation}
    \label{eq: gradient of phi}
    M(r;\mu) = \nabla_r \phi(r;\mu) = \Gamma(r) \mu .
\end{equation}

For the model $M(r;\mu)$ to fulfill \eqref{eq: div free}, the model parameters $\mu$ must be selected so that the following holds
\begin{equation}
    \nabla_r \cdot \Gamma(r) \mu = \sum_{i=x,y,z} \frac{d[\Gamma(r)\mu]_i}{dr_i} = 0 \quad \forall r \in \Omega .
\end{equation}
This constraint can be written as the linear equation system 
\begin{equation}
   \label{eq: coefficient constraints}
    D \mu = 0,
\end{equation}
where $D$ is a constant matrix derived in \cite{skog2018magnetic}.

Equations \eqref{eq: gradient of phi} and \eqref{eq: coefficient constraints} define the magnetic field model, which writes as
\begin{equation}
    M(r;\mu) = \Gamma(r) \mu, \quad D \mu =0.
\end{equation}
Finally, $M(r;\mu)$ can be reparameterized by introducing the matrix $D^{\perp} \triangleq \text{null}\{D\}$ whose columns span the null space of $D$, and then setting $\mu = D^{\perp} \theta$, where $\theta$ is a column vector of dimension equal to that of the null space of $D$. The reparameterized model is given by
\begin{equation}
\label{E: polynomial model}
    M(r;\theta) = \Phi(r)\theta,
\end{equation}
where $\Phi(r) \triangleq \Gamma(r) D^{\perp} \in \mathbb{R}^{3\times\kappa}$ is the regression matrix defined in \cite{skog2018magnetic} and $\theta \in \mathbb{R}^{\kappa}$ is the coefficient of the polynomial model; for a $l^{\text{th}}$ order polynomial the model has $\kappa$ = dim($\theta$) = $l^2+4l+3$ unknown parameters\cite{skog2021magnetic} \footnote{An example of the 1st order model used in the experiment is provided in the appendix.}. Note that the model \eqref{E: polynomial model} can be defined in either the body frame or navigation frame. Within this paper, it will be defined in the body frame. Next, a procedure for transforming the model from body frame $\alpha$ to body frame $\beta$ will be presented.

\subsection{Tranforming models between body frames}
Let the magnetic field model \eqref{E: polynomial model} be associated with the body frame, which means $M(r;\theta)$ accepts locations $r$ expressed in the current body frame and outputs magnetic field vector in the same frame. Then the magnetic field can be represented in the two body frames $\alpha$ and $\beta$, i.e., 
\begin{equation}
\begin{split}
    M^{\alpha}(r^\alpha;\theta^\alpha) &= \Phi(r^\alpha) \theta^\alpha,\\
    M^{\beta}(r^\beta;\theta^\beta) &= \Phi(r^\beta) \theta^\beta .\\
    \end{split}
\end{equation}
Here the superscripts on $M$ and $r$ denote the corresponding body frame in which they are resolved and the superscript on $\theta$ denotes the frame with which the model coefficients are associated.

\begin{figure}[tb!]
    \centering
\begin{tikzpicture}

\draw[thick,->] (0,0) -- (2.2,0) node[below] {$x$};
\draw[thick,->] (0,0) -- (0,2.2) node[left] {$y$};
\draw (0,0) node[left] {Frame $\alpha$};
\filldraw (2,1) circle (2pt) node[above right] {};
\draw[brown, thick,->] (2,1) -- (2.8, 0.7) node[right] {magnetic field vector};
\begin{scope}[shift={(-1,-1)}]
    \draw plot [smooth cycle] coordinates {(1.0,.1)(1.75,0.05)(2.5,.5) (2.9, 0.6)(3.5,2.8)(3,3.3)(2.4,3.4)(1.0,2.6)(0.3, 1.3)(0.5,0.5)} node at (2.6,0.3)[scale=1] {$\Omega$};
    \draw [fill=cyan,opacity=0.3] plot [smooth cycle] 
    coordinates {(1.0,.1)(1.75,0.05)(2.5,.5)(2.9, 0.6)(3.5,2.8)(3,3.3)(2.4,3.4)(1.0,2.6)(0.3, 1.3)(0.5,0.5)} -- cycle;
\end{scope}

\coordinate (A) at (1.9,3);
\begin{scope}[shift=(A),rotate=-15]
\draw[thick,->] (0,0) -- (2,0) node[below] {$x'$};
\draw[thick,->] (0,0) -- (0,2) node[left] {$y'$};
\draw (0,0) node[left] {Frame $\beta$};
\end{scope}
\begin{scope}[shift=(A)]
\draw[thick, dashed] (0,0) -- (2,0) ;
\draw[thick, dashed] (0,0) -- (0,2) ;
\draw[thick] (1,0) arc (0:-15:1);
\node at (1.5,-0.2) {$\Delta\phi_k$};
\end{scope}

\draw[thick,->,>=latex] (0,0) -- (A) ;
\node[text width=1cm] at (0.8,1.6)  {$\Delta p^{\alpha}$}; 

\draw[thick,->,>=latex] (0,0) -- (2,1) ;
\node[text width=1cm] at (1.2,0.7)  {$r^{\alpha}$}; 

\draw[thick,->,>=latex] (A) -- (2,1) ;
\node[text width=1cm] at (2.5,2)  {$r^{\beta}$}; 

\end{tikzpicture}
    \caption{A 2D illustration of the geometric relationship between the body frames at two consecutive times. The applicable region $\Omega$ of the magnetic field model at time $k$ is in blue, and the black dot indicates the location where the two models output the corresponding magnetic field in their coordinate frames.}
    \label{F: illustration of model propogation}
\end{figure}
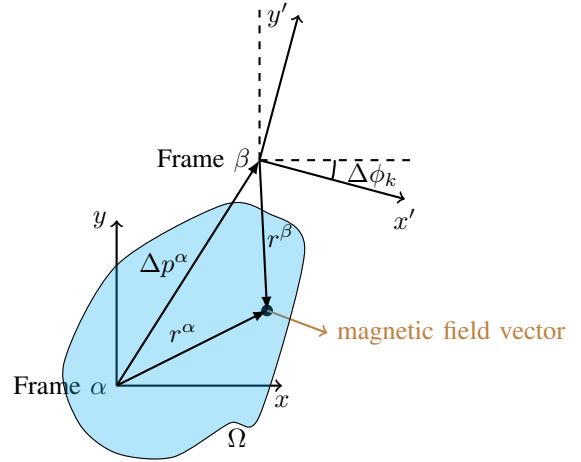

By expressing the magnetic field vector at a given location with the two models and aligning them in the same frame, the two models can be related as
\begin{subequations}
\label{eq: alpha beta magnetic field}
\begin{equation}
    M^{\beta}(r^\beta;\theta^\beta) = R_{\alpha}^{\beta} M^{\alpha}(r^\alpha;\theta^\alpha),
\end{equation}
where
\begin{equation}
    r^\alpha = (R_{\alpha}^{\beta})^{\top} r^\beta + \Delta p^{\alpha}.
\end{equation}
\end{subequations}
Here $\Delta p^{\alpha}$ denotes the translation vector expressed in body frame $\alpha$. An illustration of the geometric relationship between the two frames and the magnetic field vector is shown in Fig.~\ref{F: illustration of model propogation}.

Now consider transforming the magnetic field model from body frame $b_k$ to $b_{k+1}$. Let the relative body frame change be encoded by 
\begin{equation}
    \psi_k = \begin{bmatrix}
        \Delta p_k^{b_k} \\
        \Delta \phi_k
    \end{bmatrix},
\end{equation}
where ${\Delta p}_k^{b_k} \in \mathbb{R}^3$ and $\Delta \phi_k \in [0, 2\pi]^3$ denote the translation and orientation change from the body frame $b_k$ to $b_{k+1}$, respectively. Replacing $\alpha$ and $\beta$ with $b_k$ and $b_{k+1}$ respectively and using $\theta_k$ and $\theta_{k+1}$ to denote the corresponding coefficients, the following holds
\begin{subequations}
\label{eq: connect models}
\begin{equation}
     M^{b_{k+1}}(r^{b_{k+1}};\theta_{k+1})=R_{b_k}^{b_{k+1}}M^{b_k}(r^{b_k};\theta_{k}),
\end{equation}
where 
\begin{equation}
    r^{b_k} = (R_{b_k}^{b_{k+1}})^{\top} r^{b_{k+1}} + {\Delta p}_k^{b_k}.
\end{equation}
\end{subequations}
The rotation matrix $R_{b_k}^{b_{k+1}}$ and translation $\Delta p_k^{b_k}$  are given by
\begin{subequations}
    \begin{equation}
    R_{b_k}^{b_{k+1}} = \left(\text{exp} (\left[ \Delta \phi_k \right]_{\times})\right)^{\top},
\end{equation}
    \begin{equation}\label{eq: displacement}
    \Delta p_k^{b_k} = R_{b_k}^{n^{\top}}  \left(v_k^n T_s + (R^n_{b_k}s_k + g^n) \frac{T_s^2}{2}\right).
    \end{equation} 
\end{subequations}
Here $\Delta \phi_k = \omega_k T_s$ and $[\cdot]_{\times}$ is an operator that maps a vector in $\mathbb{R}^3$ to a skew-symmetric matrix such that $\left[a\right]_{\times} b = a \times b$.

Next, substituting the generic magnetic model in \eqref{eq: connect models} with the proposed polynomial model \eqref{E: polynomial model} yields the equality 
\begin{equation}
\label{eq: single equation relate models}
    \Phi(r^{b_{k+1}})\theta_{k+1} = R_{b_k}^{b_{k+1}}\Phi(r^{b_{k}})\theta_k .
\end{equation}

Note that for a given $\{r^{b_{k+1}}, \theta_k, \psi_k\}$, (\ref{eq: single equation relate models}) represents 3 linear equations. Since $\theta_{k+1}$ is of dimension $\kappa$, which is greater than 3, it is necessary to use more than one location vector $r$ to solve the equation system. In general,  $S=\lceil\kappa/3\rceil$ location vectors can be used to construct the equation system

\begin{subequations}
    \label{system of equations}
    \begin{equation}
    A\theta_{k+1} = B(\psi_k)\theta_k,
    \end{equation}
    where
    \begin{equation}
        A = \begin{bmatrix}
 \Phi(r^{b_{k+1}}_1)\\
 \vdots\\
 \Phi(r^{b_{k+1}}_S)\\
\end{bmatrix}, \quad B(\psi_k) = \begin{bmatrix}
R_{b_k}^{b_{k+1}}\Phi(r^{b_{k}}_1)\\
\vdots \\
R_{b_k}^{b_{k+1}}\Phi(r^{b_{k}}_S)\\
\end{bmatrix}.
    \end{equation}
\end{subequations}

If the vectors $r^{b_{k+1}}_1, \cdots, r^{b_{k+1}}_S$ can be chosen such that $A$ has full column rank, it holds that  
\begin{equation}
\label{eq: coefficient development}
\theta_{k + 1} = A^{\dagger} B(\psi_k) \theta_k,
\end{equation}
where $A^{\dagger}$ denotes the Moore–Penrose inverse of $A$.

Since the magnetic field model should describe the magnetic field locally, the update in (\ref{eq: coefficient development}), which with time implicitly expands the applicable space $\Omega$ of the model, will inevitably introduce modeling errors. To account for those errors, with a slight abuse of notation, the update of the polynomial coefficients of the magnetic field as the body frame change is modeled as
\begin{equation}
    \label{E: magnetic field subsystem}
    \theta_{k+1} = f^{\theta}(\theta_{k}, x^{\ins}_k,\tilde{u}_k, w^{\ins}_k,w^{\theta}_k),
\end{equation}
where 
\begin{equation}
     f^{\theta}(\theta_{k}, x^{\ins}_k,\tilde{u}_k,w^{\ins}_k, w^{\theta}_k) = A^{\dagger} B(\psi_k) \theta_{k} + w^\theta_k.
\end{equation}Here $w_k^{\theta}$ is assumed to be a white Gaussian noise process with zero mean and covariance matrix $\Sigma_\theta$. Note that $\psi_k$ is a function of $x^{\ins}_k$, $\tilde{u}_k$,  and $w^{\ins}_k$.

\subsection{Magnetometer Array Measurement Model}
Given the magnetic field model in \eqref{E: polynomial model}, the measurement $y_k^{(i)}\in \mathbb{R}^3$ from the $i^{\text{th}}$ sensor in the magnetometer array at time $k$ can be modeled as
\begin{equation}
\label{eq: measurement model}
    y_k^{(i)} = \Phi(r_{m_i}) \theta_k + e_k^{(i)},
\end{equation}
where $r_{m_i} \in \mathbb{R}^{3}$ denotes the location of the $i^{\text{th}}$ magnetometer in the array. Further, $e_k^{(i)}\in \mathbb{R}^{3}$ denotes the measurement error, which includes both the measurement noise and the imperfections of the magnetic-field model. The error is assumed to be white and Gaussian distributed with covariance matrix $\Sigma_{e_k^{(i)}}$. 

\subsection{Complete System}
Given the presented navigation equations and the magnetic-field model, the dynamics and observations of the full system can be described by the following state-space model. 

Let the state vector $x_k$, the process noise vector $w_k$, and the measurement noise $e_k$  be defined as 
  \begin{equation}
    x_k \triangleq \begin{bmatrix}
    x^{\ins}_k \\
    \theta_k
\end{bmatrix},  \quad      w_k \triangleq 
    \begin{bmatrix}
        w_k^{\ins}\\
         w_k^{\theta}
    \end{bmatrix}, \quad \text{and} \quad e_k \triangleq 
    \begin{bmatrix}
        e_k^{(1)}\\
        \vdots\\
        e_k^{(N)}\\
    \end{bmatrix},
  \end{equation}
respectively. Combining the models in \eqref{eq: navigation equations}, \eqref{E: magnetic field subsystem}, and (\ref{eq: measurement model}) gives the state-space model 
\begin{subequations}
\label{E: SSM}
	\begin{align}
	\label{E: SSMa}
	x_{k+1} & = f(x_k,\tilde{u}_k, w_k),\\
	y_k &= H x_k +e_k,
\end{align}
where
\begin{equation}
f(x_k,\tilde{u}_k, w_k) =\begin{bmatrix}
 f^{\ins}(x^{\ins}_k,\tilde{u}_k, w^{\ins}_k) \\
f^{\theta}(\theta_k, x^{\ins}_k,\tilde{u}_k, w^{\theta}_k)
\end{bmatrix},
\end{equation}
and
\begin{equation}
H= \begin{bmatrix}
        0_{3\times 16} & \Phi(r_{m_1})\\
        \vdots &\vdots\\
        0_{3\times 16} &\Phi(r_{m_N})\\
    \end{bmatrix}.
\end{equation}
Here, the process noise covariance $Q_k$ and the measurement noise covariance $R_k$ are
\begin{align}
     Q_k &\triangleq  \textbf{Cov}(w_k) = \text{blkdiag}(\Sigma_{a}, \Sigma_{\omega},  \Sigma_{o^{a}}, \Sigma_{o^{\omega}}, \Sigma_{\theta}), \\
    R_k &\triangleq \textbf{Cov}(e_k)= \text{blkdiag}(\Sigma_{e_k^{(1)}}, \Sigma_{e_k^{(2)}}, \cdots, \Sigma_{e_k^{(N)}}),
\end{align}
\end{subequations}
respectively.

\section{State Estimation}
\begin{figure}[tb!]
    \centering
\begin{tikzpicture}
  \node[draw, rectangle, align=center, text width=0.08\columnwidth] (box1) at (0,0) {IMU};

  \node[draw, rectangle, align=center, text width=0.4\columnwidth] (box2) at (0.35\columnwidth,0) {$\hat{x}_{k+1|k}=f(\hat{x}_{k|k},\tilde{u}_k,0)$};

  \node[draw, circle, inner sep=0pt, minimum size=1.5em] (oplus) at (0.7\columnwidth,-4) {$+$};

  \draw[->] (box2.east) -- ++(0.3\columnwidth,0) coordinate[pos=0.45] (mid) node[midway, above] {};

  \node[draw, rectangle, align=center, text width=0.3\columnwidth] (box3) at (0.35\columnwidth,-4) {Kalman filter with error state model};

  \draw[->] (box1.east) -- (box2.west) node[midway, above] {$\tilde{u}_k$};

  \draw[->] (box3.north) -- (box2.south) node[midway, left, align=left] {corrections\\$\delta \hat{x}_{k+1}$};

  \node[draw, rectangle, align=center, text width=0.25\columnwidth] (box6) at (0.7\columnwidth,-6) {Magnetometer array};

  \draw[->] (box6.north) -- (oplus.south) node[midway, right] {$+\quad y_{k+1}$};

  \draw[->] (oplus.west) -- (box3.east) node[midway, above] {};

  \node[draw, rectangle, align=center, text width=2cm] (box5) at (0.7\columnwidth,-2) {$H \hat{x}_{k+1|k}$};

  \draw (mid) -- (box5.north);

  \draw[->] (box5.south) -- (oplus.north) node[midway, right] {$-$};

\end{tikzpicture}
\caption{The flowchart of the state estimation algorithm.}
\label{F:ESKF}
\end{figure}
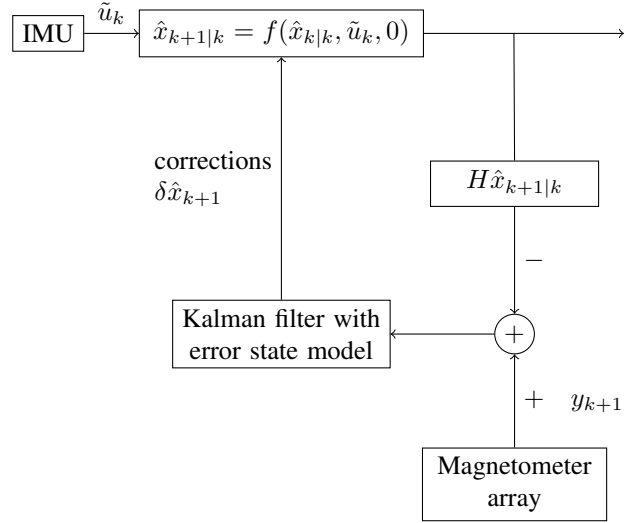

The quaternion $q_k$ in the state vector does not belong to Euclidean space. Therefore, standard nonlinear filter algorithms, such as the extended Kalman filter and unscented Kalman filter, cannot be applied to the state-space model in~(\ref{E: SSM}) without appropriate modifications. The error state Kalman filter (ESKF) presented in \cite{Joan2017Quaternion} circumvents the problem by the use of ``error quaternion'', which is a small perturbation around the ``estimated quaternion" expressed in $\mathbb{R}^3$. With reference to Fig.~\ref{F:ESKF}, the algorithm works by propagating an estimated state $\hat{x}_k$ via the state transition model in (\ref{E: SSMa}) and using a complementary Kalman filter to estimate the error state $\delta x_k$, which is then used to correct the estimated state $\hat{x}_k$. Since the ESKF is a standard algorithm, only the error propagation model that is unique to the state-space model in (\ref{E: SSM}), will here be derived. 


\subsection{Error State Definition}
The error state $\delta x_k$ is defined as
\begin{equation}
    \delta x_k \triangleq \begin{bmatrix}
        \delta x^{\ins}_k\\
        \delta \theta_k
    \end{bmatrix} \quad \text{and} \quad \delta x^{\ins}_k  = \begin{bmatrix}
    \delta p_k^n \\
    \delta v_k^n \\
    \epsilon_k\\
    \delta o_k^{a}\\
    \delta o_k^{\omega}\\
    \end{bmatrix}.
\end{equation}
Here, for the position, velocity, sensor biases, and magnetic field model parameters, the standard additive error definition is used (e.g., $\delta p_k^n = p^n_k - \hat{p}_k^n$). On the other hand, the orientation error $\epsilon_k \in \mathbb{R}^3$ satisfies the equation $R_{b_k}^n\simeq \hat{R}_{b_k}^n(I_3 + [\epsilon_k]_{\times})$. The true state $x_k$ and the estimated state $\hat{x}_k$ relate to each other via
\begin{equation}
    x_k = \hat{x}_k \oplus \delta x_k,
\end{equation}
where the operator $\oplus$ is defined by
\begin{subequations}
\begin{align}
\label{correct position}
       p^n_k &= \hat{p}_k^n + {\delta p_k^n},\\      
       v^n_k &= \hat{v}^n_k + {\delta v^n_k}, \\
       q_k &= \hat{q}_k \otimes [1\quad \frac{1}{2}\epsilon_k^{\top}]^{\top},
\\
       o_k^{a} &= \hat{o}_k^{a}  + \delta o_k^{a},\\
\label{correct angular velocity bias}
       o_k^{\omega} &= \hat{o}^{\omega}_k + 
\delta o_k^{\omega},
\\
\label{correct theta}
       {\theta}_k &= \hat{\theta}_k + 
{\delta \theta_k}.
\end{align}
\end{subequations}
\subsection{Inertial Error State Dynamics}
The dynamics of $\delta x^{\ins}_k$ has been derived in \cite{Joan2017Quaternion} and are given by
\begin{subequations}
\label{E: error state dynamics}
\begin{equation}
\delta x^{\ins}_{k+1} =F^{\ins}_k \delta x^{\ins}_k + G^{\ins}_k w^{\ins}_{k},     
\end{equation}
where
\begin{equation}
\setlength\arraycolsep{2pt}
F^{\ins}_k =\begin{bmatrix}
      I_3 & I_3 T_s & 0 & 0 & 0\\
      0& I_3 & -\hat{R}_{b_k}^n [\hat{s}_k]_{\times} T_s& -\hat{R}_{b_k}^n T_s & 0  \\
      0 & 0 & \text{exp}([\Delta \hat{\phi}_k]_{\times})^{\top} & 0 & -I_3 T_s \\
      0 & 0 & 0 & I_3 & 0\\
      0 & 0 & 0 & 0 & I_3        
    \end{bmatrix},
\end{equation}
\begin{align}
G^{\ins}_k&= \begin{bmatrix}
 0 & 0 & 0 & 0\\
 \hat{R}_{b_k}^n T_s & 0 & 0 & 0 \\
 0 & I_3 T_s & 0 & 0\\
 0 & 0 &I_3 \sqrt{T_s}& 0\\
 0 & 0 & 0    &I_3 \sqrt{T_s}
\end{bmatrix}.
\end{align}
Here, $\hat{s}_k = \tilde{s}_k - \hat{o}^{a}_k$ and $\Delta \hat{\phi}_k = (\tilde{\omega}_k - \hat{o}^{\omega}_k) T_s$.
\end{subequations}
\subsection{Magnetic Field Subsystem Error State Dynamics}
To the first order, the errors in (\ref{E: magnetic field subsystem}) propagate according to
\begin{subequations}
\begin{equation}\label{E:error propagation}
  \delta \theta_{k+1}=A^{\dagger}\begin{bmatrix}
                               B(\hat{\psi_k}) & \frac{d}{d\psi_k}\bigl(B(\hat{\psi}_k)\theta_k\bigr)
                            \end{bmatrix}
                            \begin{bmatrix}
                                           \delta \theta_k\\
                                           \delta \psi_k
                                         \end{bmatrix}
       \\ + w_k^{\theta},
\end{equation} 
where 
\begin{align}
\delta \psi_k = \psi_k - \hat{\psi}_k .
\end{align}
\end{subequations}
However, instead of expressing the error development in terms of $\delta \psi_k$, we would like to express it in terms of the orientation error $\epsilon_k$, velocity error $\delta v^n_k$, accelerometer bias estimation error $\delta o_k^{a}$, and gyroscope bias estimation error $\delta o_k^{\omega}$. To do so, note that from (\ref{eq: displacement}) we have
\begin{subequations}
\begin{equation}
\begin{split}
 \Delta \hat{p}_k^{b_k} &=  \hat{R}^{b_k}_n T_s (\hat{v}^n_k + g^n T_s/2)+  \hat{s}_k T^2_s/2 ,
\end{split}
\end{equation}
where 
\begin{equation}
     \hat{R}^{b_k}_n = (I_3-[\epsilon_k]_\times)^{-1}R^{b_k}_n ,
\end{equation}
\end{subequations}
which gives that
\begin{equation}
\begin{split}
  \delta\Delta p_k =& \Delta p_k^{b_k} - \Delta \hat{p}_k^{b_k} \\ \approx &-[\epsilon_k]_\times \hat{R}^{b_k}_n T_s(\hat{v}^n_k+g^n T_s/2)+ \hat{R}^{b_k}_n \delta v^n_k T_s\\
   =&[\underbrace{\hat{R}^{b_k}_n T_s(\hat{v}^n_k+g^n T_s/2)}_{\eta(\hat{R}^{b_k}_{n}, \hat{v}^n_k)}]_\times\epsilon_k+\hat{R}^{b_k}_n \delta v^n_k T_s .
  \end{split}
\end{equation}
Here, the second and higher-order error terms have been neglected. Moreover, it holds that
\begin{equation}
    \delta \phi_k = \Delta \phi_k - \Delta \hat{\phi}_k
    =-(\delta o_k^{\omega} + w_k^{\omega}) T_s .
\end{equation}

Bringing it all together gives the following expression for the magnetic field subsystem error state propagation 
\begin{subequations}
\label{E: polynomial coefficients error dynamics}
\begin{equation}\label{E:error propagation2}
  \delta \theta_{k+1}=F^{\theta}_k \delta \hat{x}_k + G^{\theta}_k \begin{bmatrix}
      w^{\ins}_k\\
      w^{\theta}_k
  \end{bmatrix},
\end{equation} 
where
\begin{IEEEeqnarray}{rCl}
    F^{\theta}_k &=& A^{\dagger}\begin{bmatrix}
                               B(\hat{\psi}_k) & J_1 &  J_2 
                             \end{bmatrix}  C(\hat{x}_k), 
\end{IEEEeqnarray}
\begin{equation}\label{E:L}
  G^{\theta}_k =\begin{bmatrix}
       0  \quad -A^{\dagger}J_2 T_s \quad 0 \quad 0 \quad I_{\kappa}
    \end{bmatrix}.
\end{equation}
Here,
\begin{IEEEeqnarray}{rCl}
  J_1 &=&  \frac{d}{d \Delta p_k}\bigl(B(\hat{\psi}_k)\theta_k\bigr), \\
  J_2 &=& \frac{d}{d \Delta \phi_k}\bigl(B(\hat{\psi}_k)\theta_k\bigr) ,
\end{IEEEeqnarray}
and
\begin{equation}\label{E:M}
\setlength\arraycolsep{4pt}
  \! C(\hat{x}_k)\!=\!\begin{bmatrix}
      0 & \!0 & \!0 & \!0 & \!0& \! I_\kappa\\
      0 & \! \hat{R}^{b_k}_n T_s & \! [\eta(\hat{R}^{b_k}_{n}, \hat{v}^n_k)]_{\times} & \! 0 & \! 0  & \!0 \\
      0 & \! 0 & \! 0 & \!0 & \!-I_3 T_s & \!0
    \end{bmatrix}.
\end{equation}
\end{subequations}
Combining (\ref{E: error state dynamics}) and (\ref{E: polynomial coefficients error dynamics}) gives the complete state-space model of the error state, i.e., 
\begin{subequations}
\label{eq: error state SSM}
\begin{align}
    \delta x_{k+1} &= F_k \delta x_{k} +G_kw_k, \\
    \delta y_k &= H_{\delta x}  \delta x_{k} + e_k,
\end{align}
where
\begin{equation}
     F_k = \begin{bmatrix}
         F^{\ins}_k & 0 \\
         \multicolumn{2}{c}{F^{\theta}_k}
     \end{bmatrix},\quad
    G_k = \begin{bmatrix}
         G^{\ins}_k & 0\\
         \multicolumn{2}{c} {G^{\theta}_k}
     \end{bmatrix},
\end{equation}
\begin{IEEEeqnarray}{rCl}
        H_{\delta x} = \begin{bmatrix}
        0_{3\times 15} & \Phi(r_{m_1})\\
        \vdots &\vdots\\
        0_{3\times 15} &\Phi(r_{m_N})\\
    \end{bmatrix}.
\end{IEEEeqnarray}
\end{subequations}
Here $\delta y_k \triangleq y_k- H \hat{x}_k$.

The Kalman filter can be applied to \eqref{eq: error state SSM} to estimate the error state, which is then used to correct the estimated state. The complete description of the proposed ESKF is summarized in Algorithm~\ref{alg: ESKF magINS}.
\begin{algorithm}[tb!]
\caption{ESKF algorithm}\label{alg:alg1}
\begin{algorithmic}
\renewcommand{\algorithmicrequire}{\textbf{Input:}}
\renewcommand{\algorithmicensure}{\textbf{Output:}}
\REQUIRE $\{\tilde{u}_k, y_k\}_{k=0}^N$
\ENSURE  $\{\hat{x}_{k|k}, P_{k|k}\}_{k=1}^N$
\\ \textit{Initialisation} : estimated state $\hat{x}_{0|0}$, covariance matrix $P_{0|0}$ 
\STATE {\textbf{For} $k=0$ to $N-1$ do}
\STATE \hspace{0.5cm}$ \textbf{State propagation}$
\STATE \hspace{0.5cm}$ \hat{x}_{k+1|k} \gets f(\hat{x}_{k|k},\tilde{u}_k,0) $ 
\STATE \hspace{0.5cm}$ \textbf{Error state uncertainty propagation}$
\STATE \hspace{0.5cm}$ P_{k+1|k} \gets F_k P_{k|k} F_k^{\top} + G_k Q_k G^{\top} $ 
\STATE \hspace{0.5cm}$ \textbf{Error state observation }$
\STATE \hspace{0.5cm}$ z_{k+1} \gets y_{k+1} - H \hat{x}_{k+1}$
\STATE \hspace{0.5cm}$ S_{k+1} \gets H P_{k+1|k} H^{\top} + R_{k+1}$
\STATE \hspace{0.5cm}$ K_{k+1} \gets P_{k+1|k} H^{\top} S_{k+1}^{-1}$
\STATE \hspace{0.5cm}$ \delta \hat{x}_{k+1} \gets K_{k+1} z_{k+1}$
\STATE \hspace{0.5cm}$ P_{k+1|k+1} \gets P_{k+1|k} -K_{k+1} H P_{k+1|k}$
\STATE \hspace{0.5cm}$ \textbf{Correct the estimated state}$
\STATE \hspace{0.5cm}$ \hat{x}_{k+1|k+1} \gets \hat{x}_{k+1|k} \oplus \delta \hat{x}_{k+1}$
\STATE \hspace{0.5cm}$ \textbf{ESKF reset}$
\STATE \hspace{0.5cm}$ \delta \hat{x}_{k+1} \gets 0$
\STATE {\textbf{end for}}
\end{algorithmic}
\label{alg: ESKF magINS}
\end{algorithm}

\subsection{Adaptation of the measurement noise covariance}
As previously mentioned, the polynomial magnetic field model is not perfect. The model imperfections will vary with the complexity of the magnetic field and the covariance $R_k$ should vary accordingly. One possibility to make $R_k$ adapt to the complexity of the field is to assume $R_k=\sigma_k^2 I_{3N}$ and then fit the magnetic-field model to the current observations $y_k$ and estimate $\sigma_k^2$ from the residual. That is, $\sigma_k^2$ is estimated as \cite{theodoridis2020machine},  
\begin{subequations}
      \begin{equation}
          \hat{\sigma}^2_k=\frac{1}{3N}	\lVert(I_{3N}-XX^{\dagger})y_k\rVert^2,
      \end{equation}
where $X$ is given by
       \begin{equation}
           X=\begin{bmatrix}\Phi(r_{m_1})\\
        \vdots\\
        \Phi(r_{m_N})
    \end{bmatrix}.
       \end{equation}
    
\end{subequations}

\section{Experimental Results}
To evaluate the proposed method multiple experiments were conducted using the array in Fig. \ref{F: sensor board}. In each experiment, the magnetometer array was first sitting still on the ground for a few seconds and then picked up by a person. The person then held it in his/her hands and walked in squares for a few laps before putting the board back on the ground. The true trajectory of the array was measured using a camera-based motion-tracking system. In total 8 datasets were recorded. The main characteristics of the different datasets are summarized in Table \ref{T: dataset info}.
\begin{table*}[tb!]
  \centering
\begin{threeparttable}

  \caption{Information about the datasets}
  \begin{tabular}{lllllllll}
  \hline
  \hline
   Data sequence & \textbf{LP-1} & \textbf{LP-2} & \textbf{LP-3} & \textbf{NP-1} & \textbf{NP-2} & \textbf{NP-3} & \textbf{NT-1} & \textbf{NT-2}\\
    \hline
Trajectory length\tnote{*} (m)  & 138.72 &167.07 &194.41 &136.23 &134.66 &137.76 &164.62 &137.87\\
Trajectory duration\tnote{*} (s)  & 272 &286 &332 &177 &165 &154 &185 &151\\
Average height (m)  & 0.49 &0.52 &0.55 &0.85 &0.84 &0.79 &0.73 &0.74\\
Board orientation relative to the ground   & parallel & parallel  & parallel & parallel & parallel & parallel & tilted & tilted\\    
  \hline
  \hline
  \end{tabular}
\smallskip
\scriptsize
* including the initial part of the trajectory where the position-aiding is turned on. \\
LP: low height and parallel \;  NP: normal height and parallel  \;    NT: normal height and tilted.
  \label{T: dataset info}
   \end{threeparttable}
\end{table*}

\begin{figure}[tb!]
    \centering
    \includegraphics[width=\columnwidth]{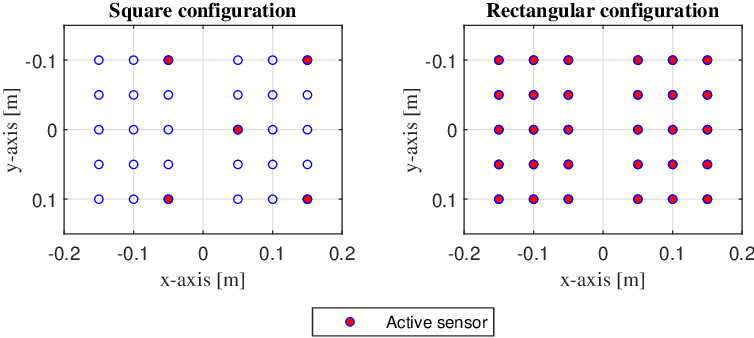}
    \caption{Sensor configurations used in the experiments. Left: Square configuration. Right: Rectangular configuration.}
    \label{F: Sensor configurations}
\end{figure}

\begin{figure}[tb!]
    \centering
    \includegraphics[width=\columnwidth]{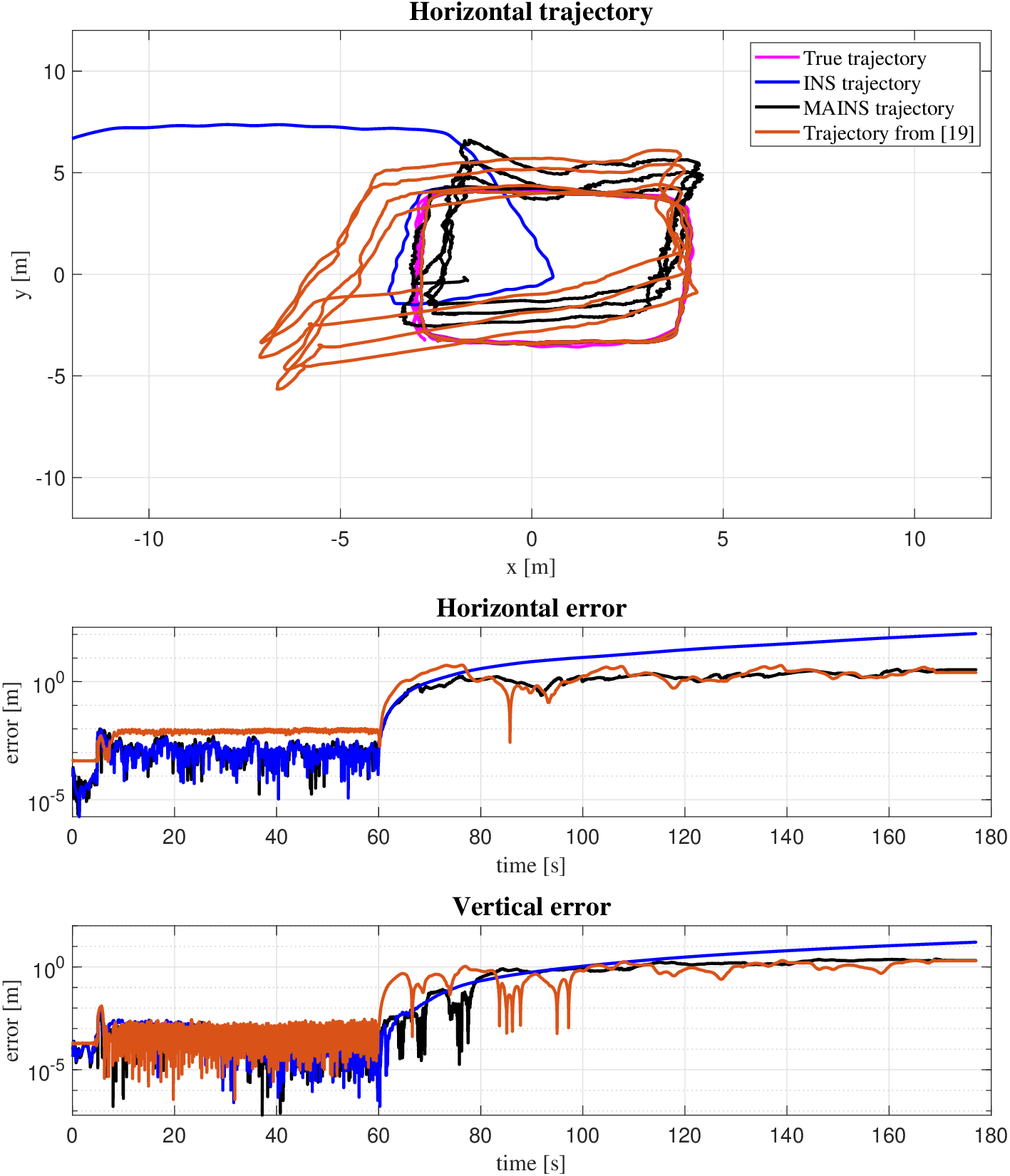}
    \caption{Estimated trajectory and the corresponding positional errors from a stand-alone INS, MAINS, and the method proposed in \cite{zmitri2020magnetic}. The square sensor configuration was used in this experiment.} 
    \label{F: Comparison of the methods}
\end{figure}

\begin{figure*}[tb!]
\centering
\includegraphics[width=.9\textwidth]{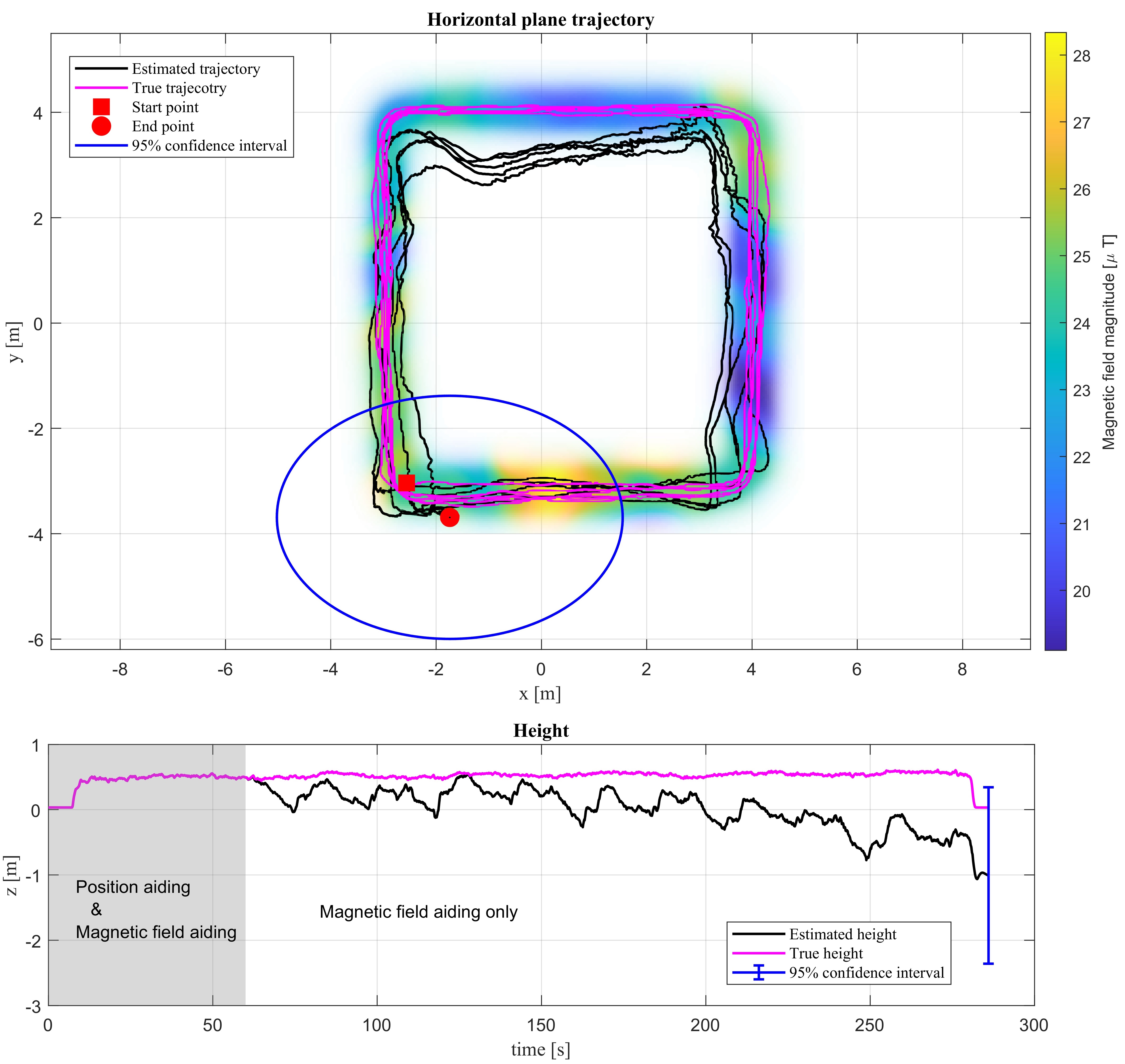}
\caption{Illustration of the estimated and the true trajectory, as well as the magnetic field magnitude along the trajectory. The magnetic field magnitude plot is created by the interpolated magnetic field measurements using a Gaussian process model.}
\label{F: Estimated trajectory and mag. field}
\label{position error}
\end{figure*}

The datasets were processed with three algorithms: a stand-alone INS; the proposed MAINS, and the method proposed in~\cite{zmitri2020magnetic}. The positions measured by the motion-tracking system were first made available to all algorithms for 60 seconds to calibrate IMU biases and stabilize state estimates. Then all systems operated without position aiding for the rest trajectory. This is similar to the scenario of a user coming into a building where GNSS signals are lost. Since the method in \cite{zmitri2020magnetic} by default is designed to use 5 magnetometers, the same sensor configuration (the left in Fig. \ref{F: Sensor configurations}) as in \cite{zmitri2020magnetic} was used when comparing the two algorithms; only the performance during the non-position-aiding part of the trajectory was evaluated. An example of the estimated trajectories estimated by the three algorithms and the corresponding positional errors are plotted in Fig. \ref{F: Comparison of the methods}. Since MAINS supports using other sensor configurations than the square configuration, the performance of the proposed algorithm was also evaluated using all the sensors in the array (see Fig. \ref{F: Sensor configurations}). An example of the trajectory estimated when using all sensors is shown in Fig. \ref{F: Estimated trajectory and mag. field}. The results, in terms of root mean square (RMS) position and velocity errors, from processing all 8 datasets with the different sensor configurations and algorithms are summarized in Table \ref{T: evaluation}. The position errors at the end of the trajectories are also shown.

It can be seen from Fig. \ref{F: Comparison of the methods} that both MAINS and the method in \cite{zmitri2020magnetic} output a trajectory with a similar shape as the true trajectory, while the INS trajectory quickly drifted away. The same conclusion can be drawn from the horizontal and vertical error plots. As expected the position error of the INS grows much faster than those of the other two methods. Looking at the results in Tab. \ref{T: evaluation}, it shows that both MAINS and the method \cite{zmitri2020magnetic} achieved superior performance in terms of horizontal error, vertical error, and speed error, compared to the stand-alone INS. However, MAINS has a consistently lower speed error than the method \cite{zmitri2020magnetic} on all datasets. Furthermore, MAINS has, in general, a lower average horizontal error, which is consistent with the observation of the trajectory shown in Fig. \ref{F: Comparison of the methods}. In terms of vertical error, the method  \cite{zmitri2020magnetic} performed poorly on the datasets where the board was tilted, while MAINS had larger errors on the datasets where the board was close to the ground. The reason for \cite{zmitri2020magnetic} performing worse is that when the board is tilted (so is the body frame), the speed errors in all three axes contributed to a larger vertical error, compared to the case when the board is flat and the vertical error comes mostly from the speed error in the z-axis. Meanwhile, the reason why MAINS produced trajectories with a large vertical drift when the board was close to the ground is that the magnetic field there was too complex for the polynomial model, which results in large fitting residuals and thus large innovations in filtering pulling the estimate away from what it should be. Comparing the performance of the MAINS algorithm with the two different sensor configurations, the benefit of using more sensors is apparent --- Vertical error was significantly reduced, and both the horizontal and vertical error at the end of the trajectory were less than 3 meters for most trajectories.

To help readers better understand the magnetic field in which the experiments were conducted and the full potential of MAINS, the trajectory estimated by MAINS with rectangular sensor configuration is plotted on top of a magnetic field magnitude plot, as shown in Fig. \ref{F: Estimated trajectory and mag. field}. It can be seen that the magnitude variance along the trajectory is around 8 $\mu$T, and the gradient varies. MAINS is capable of producing a trajectory that is very close to the true one, and more importantly, the positional error is consistently reflected by the uncertainty. One thing that may raise readers' interest is why the trajectory seems to always ``bend inwards'' near the top left corner. We cannot offer a precise explanation now, but one of the possibilities is that MAINS is sensitive to errors in magnetometer calibration parameters, which are difficult to eliminate. 

\begin{table*}[tb!]
  \centering
\begin{threeparttable}
  \caption{Summary of results from experimental evaluation}\label{T: evaluation}
  \begin{tabular}{lllllllll}
  \hline
  \hline
    & \textbf{LP-1} & \textbf{LP-2} & \textbf{LP-3} & \textbf{NP-1} & \textbf{NP-2}& \textbf{NP-3}& \textbf{NT-1} & \textbf{NT-2}\\
    \hline
Trajectory length\tnote{*} (m)   & 114.14 &139.87 &162.30 &93.68 &89.47 &86.43 &123.01 &87.26\\
Trajectory duration\tnote{*} (s) &212 &226 &272 &117 &105 &94 &125 &91\\
  \hline
  \rowcolor{lightgray}
  \multicolumn{9}{c}{Stand-alone INS}\\
  \hline
RMS Horizontal Error (m)     & 45.18 &140.04 &384.41 &43.92 &28.14 &36.0 &130.27 &36.91  \\
Horizontal Error at the end (m)  & 75.48 &346.84 &1090.47 &108.53 &77.02 &84.27 &300.99 &90.84  \\
RMS Vertical Error (m)       & 87.13 &78.45 &110.54 &6.48 &10.77 &5.08 &5.72 &10.76  \\
Vertical Error at the end (m)    & 206.12 &183.6 &256.96 &16.02 &24.73 &12.07 &11.09 &25.9  \\
RMS Speed Error (m/s)                & 1.84 &2.44 &5.96 &1.29 &1.06 &1.09 &2.86 &1.3  \\
  \hline
  \rowcolor{lightgray}
  \multicolumn{9}{c}{The method \cite{zmitri2020magnetic} with the \textbf{square} sensor configuration in Fig. \ref{F: Sensor configurations}}\\
  \hline
RMS Horizontal Error (m)      & 2.16 &2.61 &3.02 &2.38 &3.88 &1.43 &4.3 &1.18  \\
Horizontal Error at the end (m)   & 3.89 &4.42 &5.15 &2.44 &5.79 &1.56 &7.15 &0.98  \\
RMS Vertical Error (m)        & 0.96 &1.49 &2.4 &1.09 &2.03 &0.74 &12.24 &6.74  \\
Vertical Error at the end (m)      & 2.0 &2.73 &4.25 &2.0 &3.39 &0.68 &20.6 &10.96  \\
RMS Speed Error (m/s)                 & 0.35 &0.51 &0.58 &0.63 &0.51 &0.56 &0.49 &0.50  \\
  \hline
    \rowcolor{lightgray}
  \multicolumn{9}{c}{MAINS with the \textbf{square} sensor configuration in Fig. \ref{F: Sensor configurations}}\\
  \hline
RMS Horizontal Error   (m)      & 0.66 &1.29 &1.6 &1.88 &1.62 &1.89 &2.91 &0.81  \\
Horizontal Error at the end (m)   & 1.25 &1.82 &2.55 &3.22 &2.65 &2.89 &4.51 &1.04  \\
RMS Vertical Error   (m)        & 5.59 &6.32 &6.87 &1.42 &0.58 &1.16 &1.27 &0.74  \\
Vertical Error at the end(m)      & 9.61 &10.54 &11.76 &2.04 &1.21 &1.16 &2.22 &1.42 \\
RMS Speed Error (m/s)                 & 0.12 &0.17 &0.14 &0.24 &0.17 &0.29 &0.22 &0.20  \\
\hline
  \rowcolor{lightgray}
  \multicolumn{9}{c}{MAINS with the \textbf{rectangular} sensor configuration in Fig. \ref{F: Sensor configurations}}\\
  \hline
RMS Horizontal Error (m)               & 0.77 &0.84 &1.15 &2.65 &1.24 &2.32 &0.83 &1.73  \\
Horizontal Error at the end (m)    & 0.61 &1.65 &2.49 &4.21 &1.49 &3.72 &1.82 &2.38  \\
RMS Vertical Error (m)                 & 0.49 &0.57 &0.74 &2.09 &0.35 &2.11 &1.91 &0.25  \\
Vertical Error at the end (m)      & 0.56 &0.99 &1.22 &2.78 &0.89 &2.52 &3.16 &0.64  \\
RMS Speed Error (m/s)                  & 0.10 &0.11 &0.12 &0.20 &0.24 &0.24 &0.19 &0.17 \\
    \hline
  \hline
  \end{tabular}
  
\smallskip
\scriptsize
* excluding the initial part of the trajectory where the position-aiding is turned on. 
  \end{threeparttable}
\end{table*}

\section{Conclusion and Future Work}
The results presented in this paper demonstrate the effectiveness of the MAINS algorithm for magnetic-field-based indoor positioning. The proposed algorithm outperforms the stand-alone INS in terms of horizontal and vertical error, as well as speed error. Furthermore, it has a comparable performance with the
state-of-the-art method with the same sensor configuration. Having the advantage of being flexible with sensor configurations, the MAINS algorithm can, in most cases, limit the position drift to less than 3 meters after 2 minutes of navigation when using all magnetometers. 

Future work could focus on investigating loop closure detection mechanisms for magnetic field SLAM, unifying sensor calibration and parameter tuning within the MAINS framework, and exploring the impact of the magnetic field model on positioning error. Overall, the MAINS algorithm shows great promise for real-life applications of magnetic-field-based positioning.

\bibliographystyle{IEEEtran}
\bibliography{ref}

\begin{IEEEbiography}[{\includegraphics[width=1in,height=1.25in,clip,keepaspectratio]{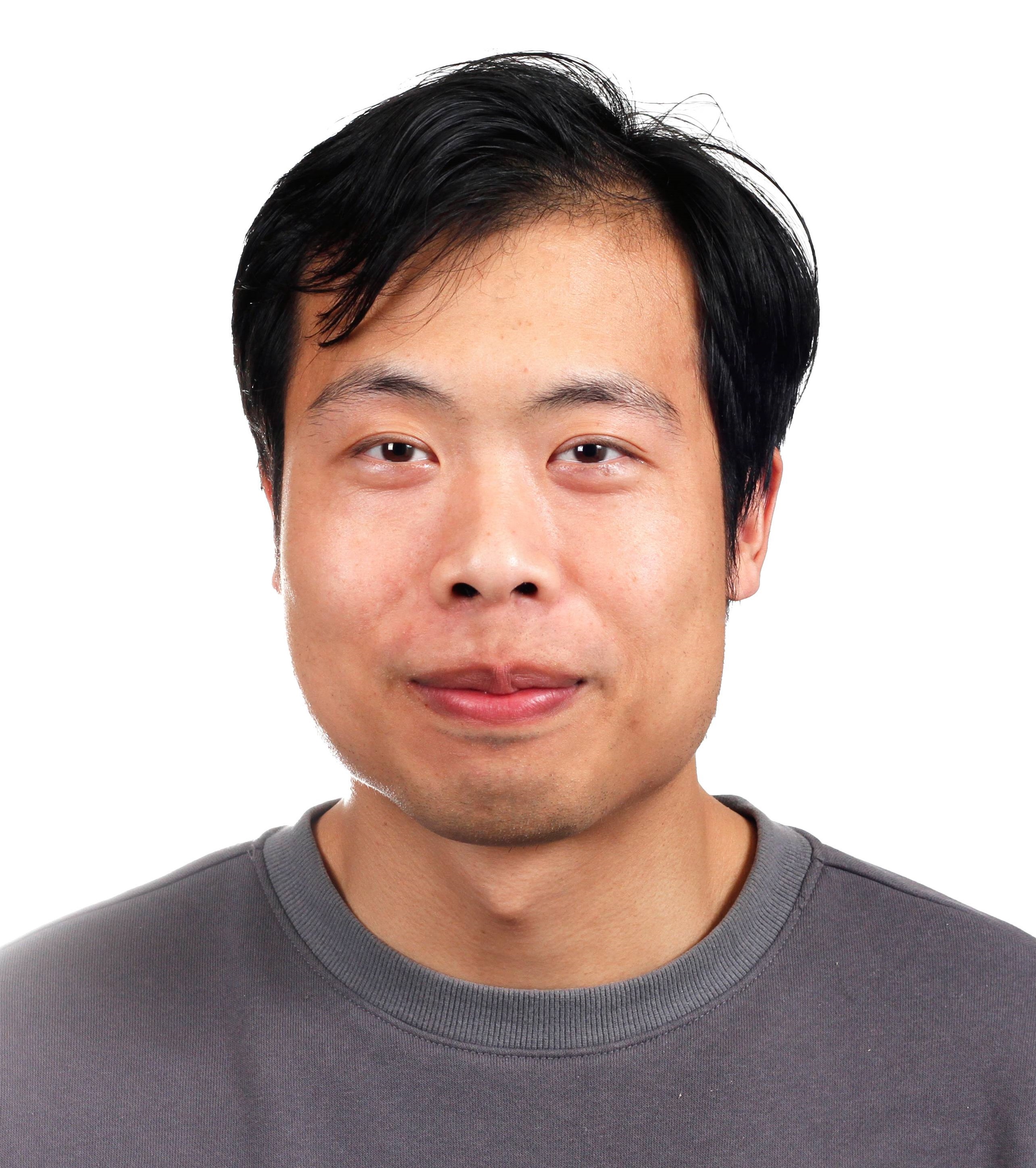}}]{Chuan Huang} (Student member, IEEE) received the B.Sc. from Beihang University in 2018 and the M.Sc. degree from China Electronics Technology Group Corporation Academy of Electronics and Information Technology in 2021. He is now a Ph.D. student at Linköping university, Sweden.

His main research interest is sensor fusion with applications to magnetic field based positioning.  
\end{IEEEbiography}

\begin{IEEEbiography}
[{\includegraphics[width=1in,height=1.25in,clip,keepaspectratio]{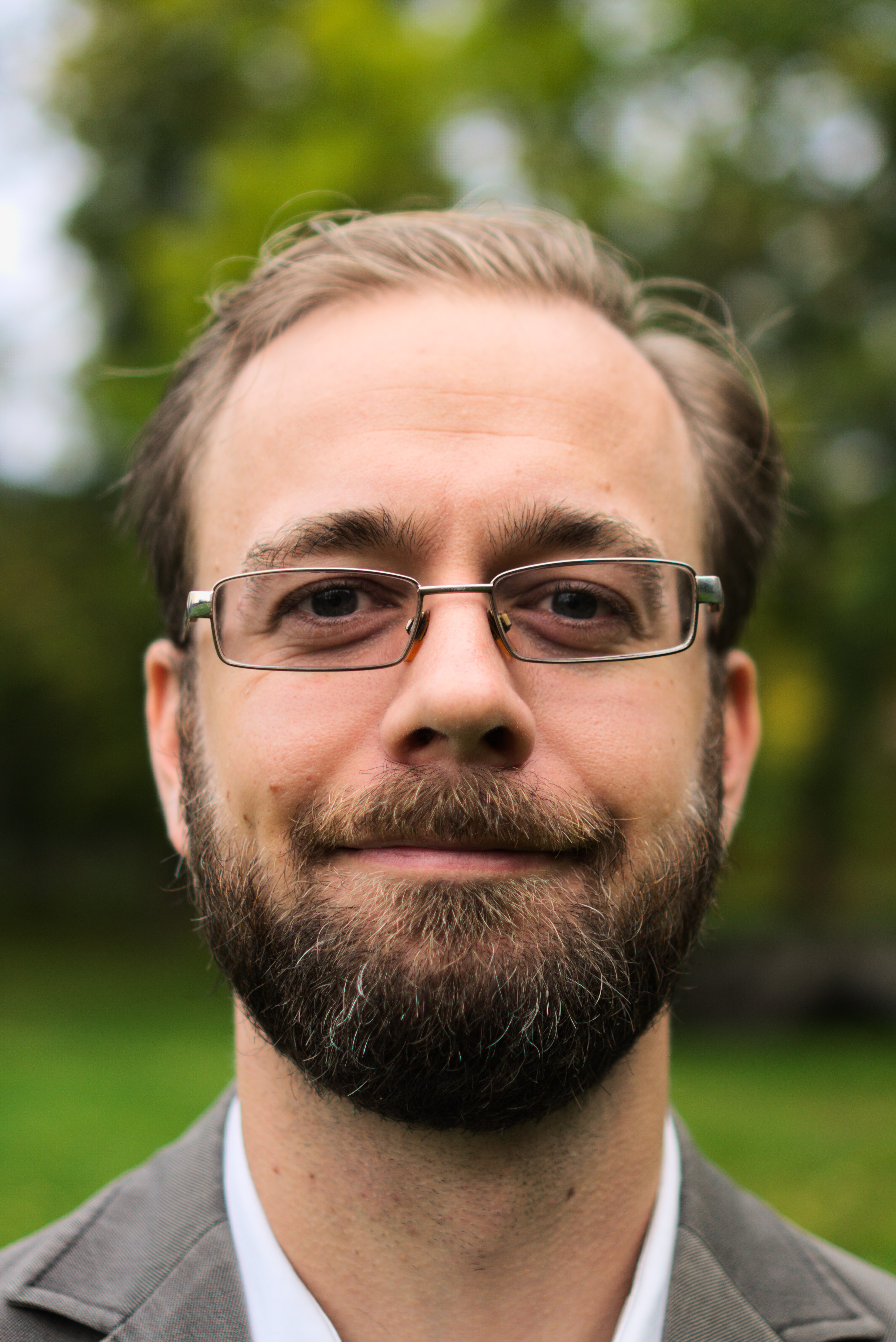}}]{Gustaf Hendeby} (Senior member, IEEE) received the M.Sc. degree in applied physics and electrical engineering in 2002 and the Ph.D. degree in automatic control in 2008, both from Linkoping University, Linkoping, Sweden. He is Associate Professor and Docent in the division of Automatic Control, Department of Electrical Engineering, Linkoping University. 

He worked as Senior Researcher at the German Research Center for Artificial Intelligence (DFKI) 2009–2011, and Senior Scientist at Swedish Defense Research Agency (FOI) and held an adjunct Associate Professor position at Linkoping University 2011–2015. His main research interests are stochastic signal processing and sensor fusion with applications to nonlinear problems, target tracking, and simultaneous localization and mapping (SLAM), and is the author of several published articles and conference papers in the area. He has experience of both theoretical analysis as well as implementation aspects. Dr. Hendeby is since 2018 an Associate Editor for IEEE Transactions on Aerospace and Electronic Systems in the area of target tracking and multisensor systems. In 2022 he served as general chair for the 25th IEEE International Conference on Information Fusion (FUSION) in Linkoping, Sweden.
\end{IEEEbiography}

\begin{IEEEbiography}[{\includegraphics[width=1in,height=1.25in,clip,keepaspectratio]{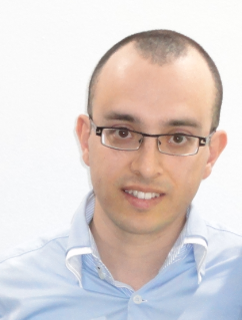}}]{Hassen Fourati} (Senior member, IEEE) received his bachelor of engineering degree in electrical engineering at the National Engineering School of Sfax (ENIS), Tunisia; master’s degree in automated systems and control at the University of Claude Bernard (UCBL), Lyon, France; and PhD degree in automatic control at the University of Strasbourg, France, in 2006, 2007, and 2010, respectively.

He is currently an associate professor of the electrical engineering and computer science at the University of Grenoble Alpes, Grenoble, France, and a member of the Dynamics and Control of Networks Team (DANCE), affiliated to the Pôle Automatique et Diagnostic (PAD) of the GIPSA-Lab. His research interests include nonlinear filtering, estimation and multisensor fusion with applications in navigation, motion analysis, mobility and traffic management. He has published several research papers in scientific journals, international conferences, book chapters and books.
\end{IEEEbiography}

\begin{IEEEbiography}[{\includegraphics[width=1in,height=1.25in,clip,keepaspectratio]{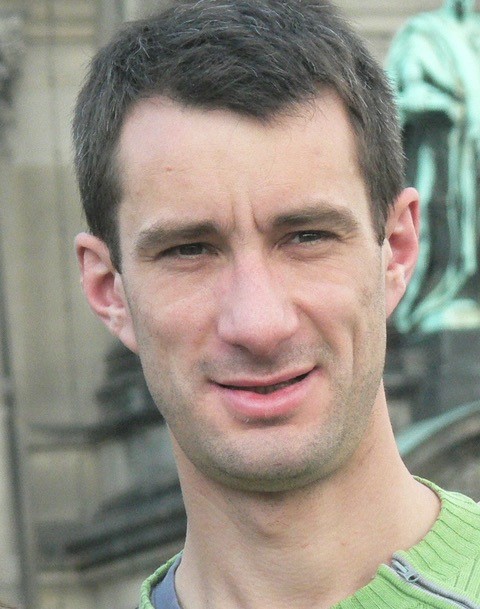}}]{Christophe Prieur} (Fellow member, IEEE) received the M.Sc. degree in mathematics from the École Normale Supérieure de Cachan, France, in 2000, and the Ph.D. degree in applied mathematics from Université Paris-Sud, Orsay, France, in 2001.

He has been a Senior Researcher with the French National Centre for Scientific Research, Paris, France, since 2011.
His current research interests include nonlinear control theory, hybrid systems, and control of partial differential equations. Dr. Prieur has been a member of the European Control Association-Conference Editorial Board (EUCA-CEB) and the IEEE-Control Systems Society (CSS) CEB. He was the Program Chair of the 9th IFAC Symposium on Nonlinear Control Systems (NOLCOS 2013) and the 14th European Control Conference (ECC 2015). He has been an Associate Editor of the IEEE Transactions on Automatic Control, the European Journal of Control, and the IEEE Transactions on Control Systems Technology. He is an Associate Editor of the Evolution Equations and Control Theory (AIMS) and the SIAM Journal on Control and Optimization. He is also a Senior Editor of the IEEE Control Systems Letters and an Editor of the IMA Journal of Mathematical Control and Information.
\end{IEEEbiography}

\begin{IEEEbiography}
[{\includegraphics[width=1in,height=1.25in,clip,keepaspectratio]{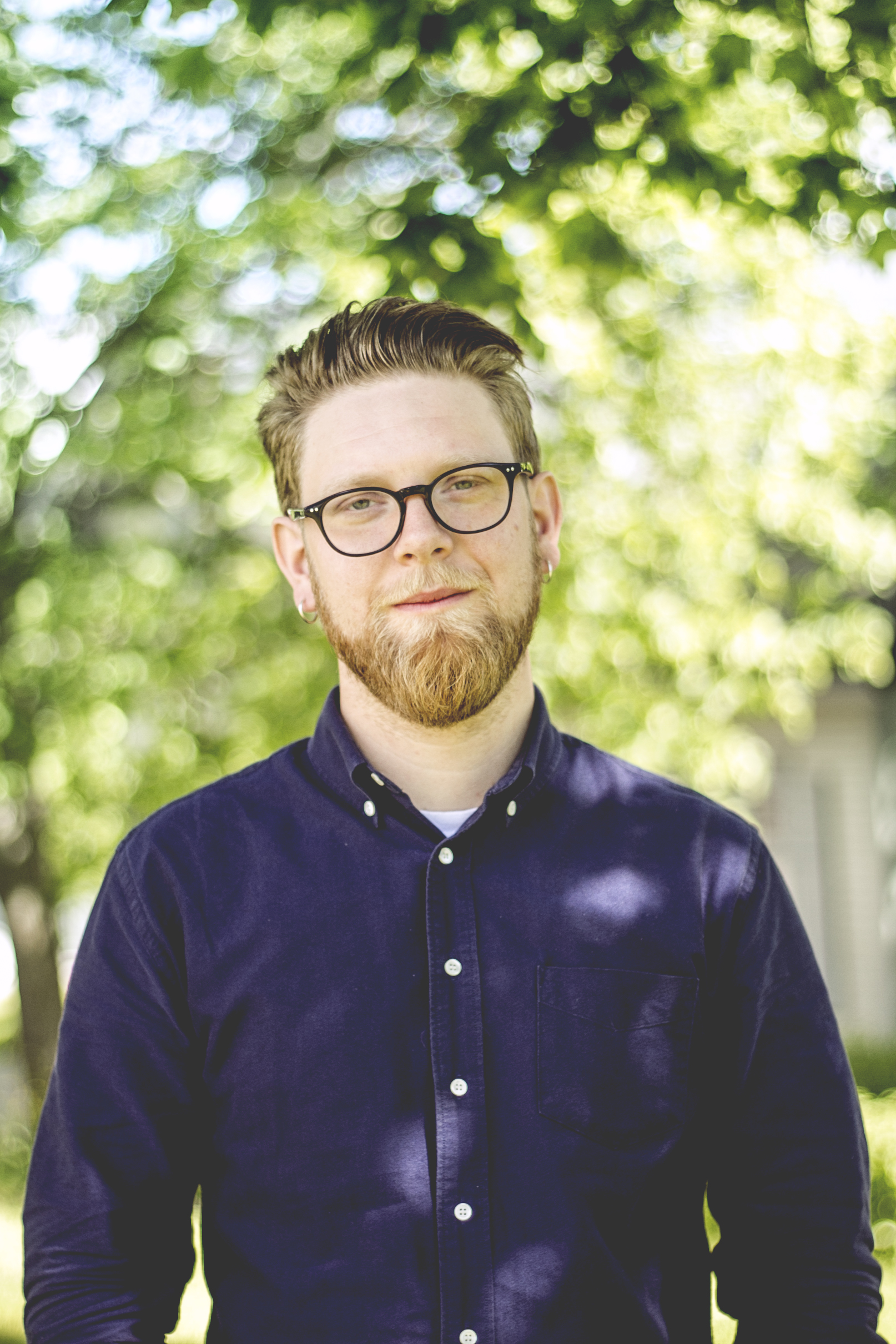}}]{Isaac Skog} (Senior Member, IEEE) received the B.Sc. and M.Sc. degrees in electrical engineering from the KTH Royal Institute of Technology, Stockholm, Sweden, in 2003 and 2005, respectively. In 2010, he received the Ph.D. degree in signal processing with a thesis on low-cost navigation systems. 

In 2009, he spent 5 months with the Mobile Multi-Sensor System Research Team, University of Calgary, Calgary, AB, Canada, as a Visiting Scholar and in 2011 he spent 4 months with the Indian Institute of Science, Bangalore, India, as a Visiting Scholar. Between 2010 and 2017, he was a Researcher with the KTH Royal Institute of Technology. He is currently an Associate Professor with Linköping University, Linköping, Sweden, and a Senior Researcher with Swedish Defence Research Agency (FOI), Stockholm, Sweden. He is the author and coauthor of more than 60 international journal and conference publications. He was the recipient of the Best Survey Paper Award by the IEEE Intelligent Transportation Systems Society in 2013.
\end{IEEEbiography}

\end{document}